\documentclass{article}

\usepackage[preprint]{neurips_2025}

\usepackage[utf8]{inputenc} % allow utf-8 input
\usepackage[T1]{fontenc}    % use 8-bit T1 fonts
\usepackage{hyperref}       % hyperlinks
\usepackage{url}            % simple URL typesetting
\usepackage{booktabs}       % professional-quality tables
\usepackage{amsfonts}       % blackboard math symbols
\usepackage{nicefrac}       % compact symbols for 1/2, etc. 
\usepackage{microtype}      % microtypography
\usepackage{xcolor}         % colors
\usepackage{amsmath}
\usepackage{microtype}
\usepackage{graphicx}
\usepackage{hyperref}
\usepackage{booktabs} % for professional tables
\usepackage{epsfig}
\usepackage{amssymb}
\usepackage{bm}
\usepackage{amsthm}
\usepackage{multirow}
\usepackage{caption}
\usepackage[super]{nth}

\usepackage{amssymb}%
\usepackage{mathrsfs}%
\usepackage{nicefrac}
\usepackage{color,xcolor}
\usepackage{pifont}% http://ctan.org/pkg/pifont
\usepackage{array}
\usepackage{wrapfig}
\usepackage{mathtools}
\usepackage{colortbl}
\usepackage{mdframed}
\usepackage{makecell}
\usepackage{subcaption}
\usepackage[ruled,noline,nofillcomment]{algorithm2e}
\usepackage{algorithmic}
\usepackage{booktabs}
\usepackage{multirow}
\usepackage{adjustbox}
\usepackage{verbatim}

\newcommand{\InCom}{\item[\textbf{Input:}]}
\newcommand{\OutCom}{\item[\textbf{Output:}]}

\title{Self-Reflective Reinforcement Learning for Diffusion-based Image Reasoning Generation}

\author{%
    Jiadong Pan$^{1}$,
    Zhiyuan Ma$^{1}$\footnotemark[2]\,\,\footnotemark[1],
    Kaiyan Zhang$^{1}$,
    Ning Ding$^{1}$,
    Bowen Zhou$^{1,2}$\footnotemark[2]\\
     $^1$ \textnormal{Department of Electronic Engineering, Tsinghua University, Beijing, China}  \\ 
     $^2$ \textnormal{Shanghai AI Laboratory, Shanghai, China}\\
       \tt\small {jiadpan@gmail.com}, {\tt\small \{mzyth,zhoubowen\}@tsinghua.edu.cn}
    }
\begin{document}

\maketitle
\renewcommand{\thefootnote}{\fnsymbol{footnote}}
\footnotetext[1]{Zhiyuan Ma leads the project.}
\footnotetext[2]{Corresponding authors.}

\begin{abstract} \label{sec:abstract}
Diffusion models have recently demonstrated exceptional performance in image generation task. 
However, existing image generation methods still significantly suffer from the dilemma of image reasoning, especially in logic-centered image generation tasks.
Inspired by the success of Chain of Thought (CoT) and Reinforcement Learning (RL) in LLMs, 
we propose SRRL, a self-reflective RL algorithm for diffusion models to achieve reasoning generation of logical images by performing reflection and iteration across generation trajectories.
The intermediate samples in the denoising process carry noise, making accurate reward evaluation difficult. To address this challenge, SRRL treats the entire denoising trajectory as a CoT step with multi-round reflective denoising process and introduces condition guided forward process, which allows for reflective iteration between CoT steps. 
Through SRRL-based iterative diffusion training, we introduce image reasoning through CoT into generation tasks adhering to physical laws and unconventional physical phenomena for the first time.
Notably, experimental results of case study exhibit that the superior performance of our SRRL algorithm even compared with GPT-4o. 
The project page is \url{https://jadenpan0.github.io/srrl.github.io/}.
  %By fine-tuning Stable Diffusion with SRRL algorithm, we achieve image reasoning related to physical laws and unconventional physical phenomena. 
  %Impressively, reasoning generated images of SRRL not only rival but also surpass those of GPT-4o. 
  
\end{abstract}
\vspace{-10pt}
\section{Introduction}\label{sec:introduction}
\vspace{-4pt}
% Recently, the field of text-to-image (T2I) generation is developing rapidly~\cite{ho2020denoising, ramesh2021zero, ramesh2022hierarchical}, aiming to generate images that adhering to textual descriptions. Diffusion models have demonstrated powerful capabilities in image generation tasks~\cite{saharia2022photorealistic, ramesh2022hierarchical,podell2023sdxl,ma2024safe,zhang2023adding,ruiz2023dreambooth,ma2024adapedit}, which introduces denoising process transforming Gaussian noise into images. Training on large datasets has significantly improved the generation quality of T2I diffusion models~\cite{rombach2022high, ding2021cogview,saharia2022photorealistic}. However, the models often face text-image misalignment issues when generating images with complex textual descriptions~\cite{jiang2024comat,huang2023t2i}. 

Recent years have witnessed the remarkable success of text-to-image (T2I) models~\cite{ho2020denoising, ramesh2021zero, ramesh2022hierarchical,ma2024neural}. As the pioneering model among many T2I models, diffusion models have demonstrated powerful abilities in generating realistic images~\cite{saharia2022photorealistic, ramesh2022hierarchical,podell2023sdxl,ma2024safe,zhang2023adding,ruiz2023dreambooth,ma2024adapedit}. Existing works introduce ControlNet~\cite{zhang2023adding} and T2I-Adapter~\cite{mou2024t2i} to enhance the controllability of image generation. However, these models still lack the ability of reflective reasoning, resulting in issues that images do not adhere to physical laws, where images may be visually stunning but logically inconsistent~\cite{jiang2024comat,huang2023t2i}.

%We also notice that diffusion models lack the ability for reflective thinking, which results in insufficient capabilities of reasoning and imagination.

%To alleviate this dilemma, r
Reinforcement learning (RL) based training methods~\cite{black2023training, wallace2024diffusion, fan2023dpok, hu2025towards, majumder2024tango}, including Direct Preference Optimization (DPO) and Proximal Policy Optimization (PPO), have recently been integrated into diffusion models to enhance specific capabilities, such as text-image alignment and human feedback alignment. 
%By considering the step-by-step denoising process as a multi-step decision-making process~\cite{black2023training}, RL optimizes the parameters by evaluating the entire trajectory generated by diffusion models.
DPO aligns diffusion models to human preferences by directly optimizing on comparison data, relying on high-quality user feedback, which leads to high collection costs. 
PPO optimizes the parameters by considering the step-by-step denoising process as a multi-step decision-making process~\cite{black2023training}, which 
treats noisy samples at each timestep as states, denoising process at each timestep as actions, and evaluated score of the final images as rewards. 
%PPO in diffusion models samples generation trajectories and updates model parameters based on the state, action, and reward defined above.
However, PPO optimizes the entire trajectory according to the final images by outcome reward models (ORMs), lacking the ability for reflective reasoning, which results in insufficient capabilities of complex logical image generation.
%reasoning and imagination.
%However, RL algorithms in diffusion models~\cite{} often lack reflection capability, which optimizes the entire trajectory and cannot dynamically adjust the generation of the next trajectory based on the feedback from the generated images. 

Reflective reasoning through Chain-of-thought (CoT)~\cite{wei2022chain} has been widely explored in LLMs by allowing models to decompose complex problems into several intermediate reasoning steps~\cite{guo2025deepseek, jaech2024openai, hui2024qwen2,lightman2023let}. 
Despite CoT being widely used in LLMs to increase the ability of solving complex NLP problems, there is relatively less work~\cite{guo2025can,jiang2025t2i} on enhancing reasoning capabilities in the field of image generation. Very recently, some works~\cite{guo2025can,jiang2025t2i} explore CoT in auto-regressive image generation architecture. However, there remains a significant challenge, which is exploring introducing CoT into diffusion models to enhance image reasoning capabilities.
The step-by-step denoising process of diffusion models produces noisy intermediate samples that are difficult to evaluate, thereby hindering the implementation of CoT reasoning during the denoising process.
%Diffusion models obtain the final results through multiple denoising steps, making the intermediate samples noisy and challenging to evaluate, which makes it difficult to use CoT in the denoising process of diffusion models. 
%To address the issue, we explore to obtain 

% Chain-of-thought (CoT)~\cite{wei2022chain} has been widely used for facilitating the reasoning ability of large language models (LLMs) by allowing models to decompose complex problems into several intermediate steps. CoT provides LLMs with test-time scaling ability~\cite{muennighoff2025s1}, enabling the models to solve complex reasoning problems (such as code and math problems)~\cite{}, and allowing the models to self-reflect and correct answers.~\cite{} Despite CoT being widely used in the natural language processing field and the increasing ability of LLMs to solve complex problems~\cite{}, there is relatively less work~\cite{guo2025can} on enhancing model reasoning capabilities in the field of image generation. Some works~\cite{guo2025can} explored CoT in auto-regressive image generation architecture, and a current challenge in the T2I field is exploring the reasoning capabilities of diffusion models.

\begin{figure}
    \centering
    \includegraphics[width=1\textwidth]{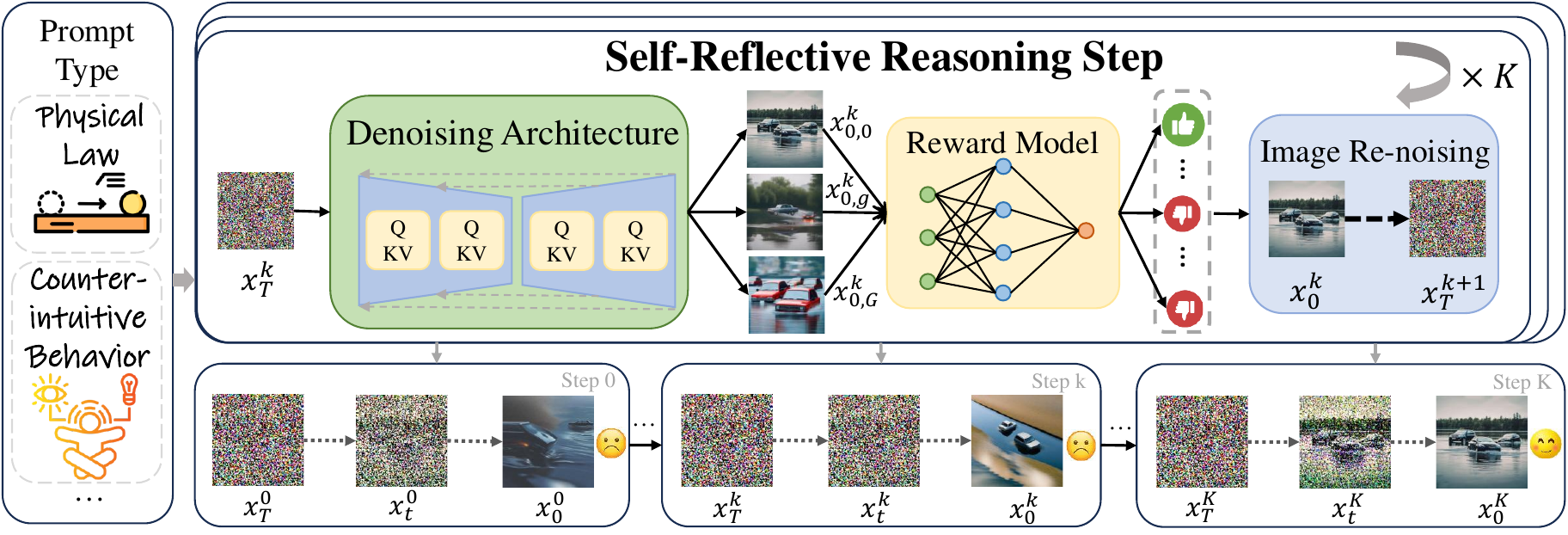}
    \vspace{-16pt}
    \caption{
    Illustration of self-reflective reasoning step. Through self-reflective processes of repeated denoising and re-noising, diffusion models achieve image reasoning generation adhered to physical laws and counterintuitive physical phenomena.
    } 
    \label{fig:prefix}
\vspace{-16pt}
\end{figure}

In this paper, we present a novel self-reflective RL algorithm \textbf{SRRL} of diffusion models, introducing CoT into diffusion models to provide self-reflective capabilities by RL training to achieve image reasoning generation.
%In this paper, we aim to incorporate CoT into diffusion models to provide self-reflective ability. We present a novel self-reflective RL algorithm \textbf{SRRL} of diffusion models, enabling them to obtain self-reflective capabilities during RL training process to achieve image reasoning generation. 
Specifically, SRRL incorporates multi-round reflective denoising process and condition guided forward process, treating the entire diffusion denoising trajectory as a step and constructing CoT between different trajectories instead of in a single denoising trajectory, which avoids the challenges of predicting rewards of noisy samples. 
Illustration of self-reflective reasoning step is shown in Fig.~\ref{fig:prefix}.
With self-reflective capabilities, SRRL achieves image reasoning generation—for instance, ensuring that generated images adhere to physical laws, such as depicting plants growing taller with sunlight compared to those without in Fig.~\ref{fig:physic}. Experimental results demonstrate that diffusion models trained by SRRL can generate images adhering to physical laws and counterintuitive scenarios. More impressively, images adhering to physical laws and counterintuitive physical phenomena generated through self-reflective reasoning of SRRL rival or surpass those generated by GPT-4o~\cite{hurst2024gpt}.

% In this paper, we present a novel self-reflective RL algorithm \textbf{SRRL} of diffusion models, which enables models to obtain self-reflection capabilities during RL training process, and enhances its reasoning ability through a self-feedback mechanism during inference process. Specifically, during the training process, SRRL provides the model's ability to perform multi-round self-feedback, improving its capacity to adjust the generation trajectory during the diffusion denoising process based on textual descriptions. During the inference process, the model utilizes the self-feedback capabilities learned during the training process to denoise noisy latents, and employs classifier-free guidance~\cite{ho2022classifier} to inject text conditions in forward process of diffusion models, allowing the model to enhance its reasoning ability through multiple rounds of self-feedback. 
% Through SRRL, experimental results.

Our contributions can be summarized as:
\begin{itemize}
    \item We introduce a self-reflective RL algorithm SRRL, enabling diffusion models with the ability for self-reflective thinking and imagination.
    %\item We explore incorporating CoT into the diffusion denoising process, allowing for reasoning time scaling, enabling diffusion models to generate images with complex reasoning content.
    \item We explore introducing CoT into the generation process of diffusion models, allowing process reward models (PRMs) to address the issue of diffusion models being unable to self-reflect based on noisy intermediate results.
    \item Experimental results indicate that SRRL achieves image reasoning generation adhering to both physical laws and counterintuitive physical phenomena. Specifically, experimental samples of SRRL exhibit superior quality even compared to GPT-4o.
    %can rival or surpass those generated by GPT-4o.
\end{itemize}

%% lack image cot

%% CFG can inject text condion

%% 设计了

%Diffusion models achieve the transformation from a Gaussian distribution to the real image distribution by learning the marginal distribution at each step of the denoising process, which makes 

\vspace{-10pt}
\section{Related Work}
\vspace{-4pt}

\subsection{Text-to-image Diffusion Models}
\vspace{-4pt}

Diffusion models are widely used in text-to-image (T2I) tasks due to their exceptional performance in generating high-quality images~\cite{song2020score, dhariwal2021diffusion,nichol2021improved,song2020improved,ma2025efficient}. Diffusion models generate images by denoising noisy images under the guidance of the text conditions. Many works, such as Stable Diffusion~\cite{rombach2022high}, Imagen~\cite{saharia2022photorealistic}, DALL-E~\cite{ramesh2021zero}, GPT-4o~\cite{hurst2024gpt}, demonstrate the ability of diffusion models in T2I tasks. 
The alignment between text and images has become an important metric for improving the effectiveness of the model.
Classifier-free guidance (CFG)~\cite{ho2022classifier} is introduced into diffusion models to enhance text conditions and text-image alignment. Some works~\cite{chung2024cfg++,bradley2024classifier,li2024self} improve the generation quality and text-image alignment by optimizing CFG. Zigzag diffusion sampling
~\cite{bai2024zigzag} incorporates a self-reflection mechanism leveraging CFG to accumulate semantic information during inference process.
However, they do not consider allowing models to learn reasoning, which leads to their inability to generate logical images adhering to physical laws.

\subsection{Reinforcement Learning of Diffusion Models}

%% DDPO ,DPOK ,B2 RL

%Reinforcement learning (RL) is a kind of methods for solving sequential decision making tasks by designing agents to interact with environment to maximize accumulated reward. 
Reinforcement Learning from Human Feedback (RLHF)~\cite{ouyang2022training} is employed for better alignment of diffusion models to human preferences. Some reward models~\cite{hessel2021clipscore, xu2023imagereward, lin2024evaluating} are trained to enhance aesthetic quality, text-image alignment, and so on, to align with human preferences. 
Diffusion denoising process can be seen as a sequential decision-making problem~\cite{black2023training}, allowing the application of RL algorithms~\cite{black2023training, fan2023dpok, hu2025towards,ren2024diffusion}. DDPO~\cite{black2023training} is a policy gradient algorithm treating diffusion denoising process as Markov decision process and using proximal policy optimization (PPO)~\cite{schulman2017proximal} updates. However, these algorithms use outcome reward models (ORMs) due to the challenge of evaluating intermediate noisy images and cannot self-reflective reasoning based on a single denoising process. 

%Reinforcement Learning from Human Feedback (RLHF) is employed for better aligment of diffusion models to human preferences. Some reward models~\cite{} are trained to enhance aesthetic quality, text-image alignment, and so on, to align with human preferences. 

%% CLIP score, image reward, vqascore%%CLIP评分、图像奖励、vqascore
%\subsection{Generative Model Reasoning}
\subsection{Reflective Reasoning Through Chain-of-Thought}

Large language models (LLMs) and multi-modal large language models (MLLMs) are discovered to simulate human thought process by reflective reasoning based on their understanding and generation skills~\cite{meng2025mm, yang2025r1,jiang2025mme}. Recent works~\cite{jaech2024openai,guo2025deepseek,wei2022chain} incorporate Chain-of-Thought (CoT) to achieve superior performance in text generation tasks, such as mathematics~\cite{zhang2024mavis,lu2023mathvista}, coding~\cite{jain2024livecodebench}, and image understanding~\cite{huang2025vision} problems. On the contrary, the exploration of CoT in image generation has been more limited. Some works~\cite{guo2025can,jiang2025t2i} explore incorporating CoT in image generation tasks. However, it uses the auto-regressive architecture as the backbone, without exploring the potential of CoT in T2I diffusion models, which are more widely used in commercial applications.

\vspace{-8pt}
\section{Method} \label{sec:method}
\vspace{-4pt}

\begin{figure}
    \centering
    \includegraphics[width=1\textwidth]{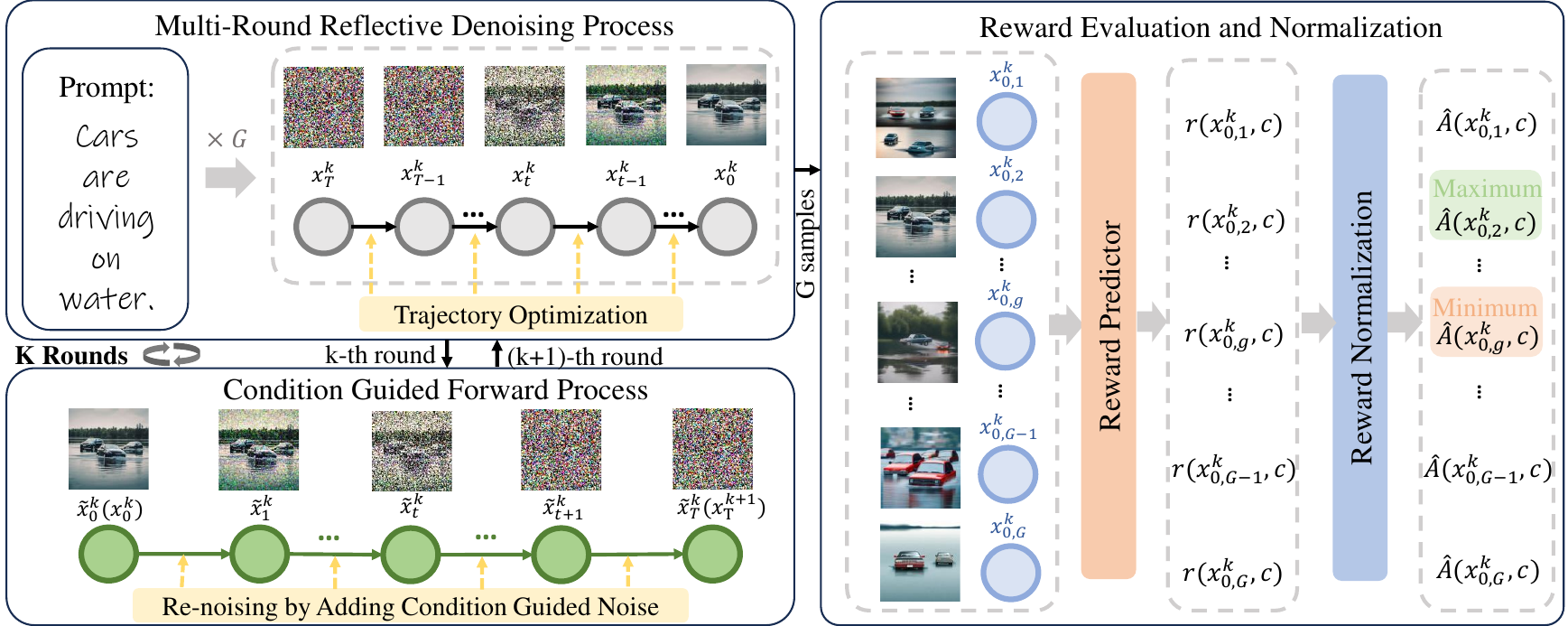}
    \vspace{-16pt}
    \caption{
    Overview of SRRL. SRRL includes two processes: multi-round reflective denoising process and condition guided forward process. These two processes are repeated for $K$ rounds.
    } 
    \label{fig:method}
\vspace{-16pt}
\end{figure}

In this section, we first introduce the training of diffusion models using reinforcement learning (RL) algorithms and self-reflective RL algorithm of diffusion models SRRL in Sec.~\ref{method:problem_formulation}. 
Then we propose multi-round reflective denoising process in Sec.~\ref{method:multiround_rl} and condition guided forward process in Sec.~\ref{method:condition_forward_process}. These two processes together constitute SRRL algorithm, which is illustrated in Fig.~\ref{fig:method}.

\vspace{-3pt}
\subsection{Problem Formulation}\label{method:problem_formulation}
\vspace{-3pt}

\subsubsection{Reinforcement Learning Training of Diffusion Models}
\vspace{-2pt}

Text-to-image diffusion models generate images by progressively denoising noisy images. 
We follow the formulation of diffusion models in denoising diffusion probabilistic models (DDPMs)~\cite{ho2020denoising}.
Diffusion models are composed of two processes: forward process and denoising process.

\textbf{Forward Process}. Given a dataset with samples $x_0\sim q_0(x_0|c)$ where $q_0$ is the data distribution and corresponding to text condition $c$, forward process is a Markov chain that gradually adds Gaussian noise into $x_0$ in T timesteps according to the variance schedule $\beta_t$:
\vspace{-8pt}
\begin{equation}
    q(x_t \vert x_{t-1}) = \mathcal{N}(x_t; \sqrt{1 - \beta_t} x_{t-1}, \beta_t\mathbf{I}), \quad \quad q(x_{1:T}|x_0)= \prod_{t=1}^T q(x_t|x_{t-1})
\end{equation}
Forward process constructs an approximate posterior distribution, and the goal of denoising process is to approximate it.

\textbf{Denoising Process} is a Markov chain, which can be seen as a Markov decision process (MDP).
\vspace{-6pt}
\begin{equation}
    p_\theta(x_{t-1} \vert x_t,c) = \mathcal{N}(x_{t-1}; \boldsymbol{\mu}_\theta(x_t,c, t), \boldsymbol{\Sigma}_t), \quad  \quad p_\theta(x_{0:T} \vert c)= p(x_T)\prod_{t=1}^T p_\theta(x_{t-1} \vert x_t,c)
\end{equation}
\vspace{-6pt}
where $\boldsymbol{\mu}_\theta(x_t,c,t)$ is predicted by a diffusion model $\theta$, and $\Sigma_t$ is variance related to timestep $t$. 

Given samples $x_0$ and text condition $c\sim p(c)$, text-to-image diffusion models generate images according to text condition $c$. 
Classifier-free guidance (CFG)~\cite{ho2022classifier} enhances text conditions to improve image generation quality by subtracting the predicted unconditional noise from the conditional noise:
\vspace{-8pt}
\begin{equation}
    \tilde\epsilon_\theta(x_t,c,t,\lambda)= \epsilon_\theta(x_t,c,t)+\lambda(\epsilon_\theta(x_t,c,t)-\epsilon(x_t,\phi,t))
\end{equation}
Here $\epsilon_\theta(x_t,c,t)$ is the conditional noise satisfying $\mu_\theta (x_t,t,c)=  \frac{1}{\sqrt{\alpha_t}}(x_t- \frac{\beta_t}{\sqrt{1-\bar{\alpha_t}}}\epsilon_\theta(x_t,c,t))$, where $\alpha_t=1-\beta_t$,$\bar{\alpha_t}=\prod_{i=1}^t \alpha_i$~\cite{ho2020denoising}. $\phi$ refers to no condition during the denoising process. %\cite{bai2024zigzag} explores to leverage the guidance gap of CFG to inject text condition  .

The goal of DDPMs is approximating $q_0(x_0|c)$ with 
$p_\theta(x_0|c)= \int p_\theta(x_{0:T}|c) dx_{1:T}$. The denoising process can be seen as a multi-step MDP $\tau= (s_T,a_T,s_{T-1},a_{T-1},\cdots,s_0,a_0)$:

\vspace{-4pt}
\resizebox{1\hsize}{!}{$
%\begin{equation}
    s_t= (c,t,x_t), \quad \quad a_t= x_{t-1}, \quad \quad  \pi_\theta(a_t|s_t)=p_\theta(x_{t-1}|x_t,c), \quad \quad R(s_t,a_t)= \begin{cases}
r(x_0, c), & \text{if } t = 0 \\
0, & \text{otherwise}
\end{cases}
%\end{equation}
$}
where $s_t$ is the state at each timestep, $a_t$ is the action to denoise $x_t$ to $x_{t-1}$, $\pi_\theta$ defines the action selection strategy, and $R$ is the reward, which is given by models or human preferences. Therefore, the denoising process of diffusion models can be viewed as an RL task in which diffusion models act as agents to make decisions (denoising process). The goal of RL is to maximize the expected cumulative reward over the diffusion denoising trajectories sampled from the policy, which can be formulated as:
\begin{equation}
    \mathcal{J}_{RL}(\theta)= \mathbb{E}_{c\sim p(c),x_0\sim p_\theta(x_0 \vert c)}[r(x_0,c)]
\end{equation}
where $p(c)$ is the distribution of text descriptions of images. 

% \subsubsection{Conditional Generation of Text-to-image Diffusion Models}

% Given samples $x_0$ and text condition $c\sim p(c)$, text-to-image diffusion models generate images according to text condition $c$. 
% Classifier-free guidance (CFG)~\cite{ho2022classifier} enhances text conditions to improve image generation quality by subtracting the predicted unconditional noise from the conditional noise:
% \begin{equation}
%     \tilde\epsilon_\theta(x_t,c,t,\lambda)= \epsilon_\theta(x_t,c,t)+\lambda(\epsilon_\theta(x_t,c,t)-\epsilon(x_t,\phi,t))
% \end{equation}
% Here $\epsilon_\theta(x_t,c,t)$ is the conditional noise satisfying $\mu_\theta (x_t,t,c)=  \frac{1}{\sqrt{\alpha_t}}(x_t- \frac{\beta_t}{\sqrt{1-\bar{\alpha_t}}}\epsilon_\theta(x_t,c,t))$, where $\alpha_t=1-\beta_t$,$\bar{\alpha_t}=\prod_{i=1}^t \alpha_i$~\cite{}. $\phi$ refers to no condition during the denoising process.
\vspace{-3pt}
\subsubsection{Self-Reflective Reinforcement Learning}
\vspace{-3pt}

%Previous RL algorithms~\cite{black2023training, fan2023dpok, hu2025towards} only optimize a single denoising trajectory, 
Existing reinforcement learning algorithms \cite{black2023training, fan2023dpok, hu2025towards} optimize only a single denoising trajectory and can only utilize outcome reward models (ORMs) without reflective reasoning capabilities.
%without considering the empowerment of the model's self-reflection capability through multiple rounds of denoising trajectory optimization. 
Different from them, SRRL aims to optimize the cumulative denoising trajectory, enabling it to utilize process reward models (PRMs) from intermediate results, which enables self-reflective reasoning process. The objective of SRRL is:
\begin{equation}
    \mathcal{J}_{SRRL}(\theta)= \mathbb{E}_{c\sim p(c), x_0\sim p_\theta(x_0|c), k\sim U(0,K)}[r(x_{0}^k,c)]
\end{equation}
where $k$ refers to the $k$-th iteration of the reflection process, $x_{0}^k$ refers to the $k$-th intermediate sample for evaluation, $U$ refers to uniform distribution. SRRL includes multi-round reflective denoising process and condition guided forward process, which will be detailed in the following sections.

\vspace{-3pt}
\subsection{Multi-Round Reflective Denoising Process}\label{method:multiround_rl}
\vspace{-3pt}

% challenge  ,SRRL 
Diffusion models suffer from the issue that 
reward prediction is limited to final images, preventing the introduction of PRMs and resulting in a lack of reflection capability. To address the issue, SRRL incorporates multiple rounds of RL optimization in the denoising process, providing PRMs and aiming to endow the model with self-reflection capability.
Specifically, after each round of the denoising process, SRRL evaluates intermediate images using reward models, which provide process rewards for the entire multi-round process. In the subsequent rounds, SRRL optimizes the trajectory based on the intermediate rewards from previous rounds.

SRRL leverages policy gradient estimation by computing likelihoods and gradients of likelihoods:
\begin{equation}
    \nabla_\theta \mathcal{J}_{SRRL} = \mathbb{E}_{c\sim p(c), x_0\sim p_\theta(x_0|c),k\sim U(0,K)} [\sum_{t=0}^{T_k} \nabla_\theta \log p_\theta (x_{t-1}^k|x_{t}^k,c)r(x_{0}^k,c)]
\label{equ:policy_gradient}
\end{equation}

Evaluation of the above requires sampling from the multi-round denoising process, which can be seen as a long MDP $\tau_{SRRL}=(s_T^0, a_T^0,\cdots, s_0^0,a_0^0,\cdots,s_T^k, a_T^k,\cdots, s_0^k,a_0^k, \cdots, s_0^K,a_0^K )$. The reward includes process rewards of intermediate samples: $R(s_t^k,a_t^k)= \begin{cases} r(x_0^k, c), & \text{if } t = 0 \\
0, & \text{otherwise} \end{cases}$.

We apply Proximal Policy Optimization (PPO)~\cite{schulman2017proximal} algorithm, including importance sampling and clipping. Besides, we use reward normalization and remove the value function, similar to Group Relative Policy Optimization~\cite{shao2024deepseekmath} algorithm, and contrastive sampling~\cite{mikolov2013distributed} is introduced. The PPO update objective is:

%\begin{equation}
\resizebox{1\hsize}{!}{$
    \nabla_\theta\mathbb{E}_{c\sim p(c),k\sim U(0,K)} \frac{1}{G_c}\sum_{i=1}^{G_c} \left(  \sum_{t=1}^T[\min( \frac{p_\theta(x_{t-1}^k|x_t^k,c)}{p_{old}(x_{t-1}^k|x_t^k,c)}\hat{A}_i^k, \text{clip}(\frac{p_\theta(x_{t-1}^k|x_t^k,c)}{p_{old}(x_{t-1}^k|x_t^k,c)}, 1-\epsilon,1+\epsilon)\hat{A}_i^k)\right)
$}

where $G_c$ is the number of remaining samples after contrastive sampling (selecting the maximum and minimum reward values). $\hat{A_i^k}$ is calculated through reward normalization: $\hat{A}_i^k=\frac{r(x_{0,i}^k,c)-\text{mean}(\{r(x_{0,1}^k,c),\cdots,r(x_{0,G}^k,c)\})}{\text{std}(\{r(x_{0,1}^k,c),\cdots r(x_{0,G}^k,c)\})}$, where $G$ is the number of samples before contrastive sampling and $k$ is the $k$-th reflection round.

\vspace{-5pt}
\subsection{Condition Guided Forward Process}\label{method:condition_forward_process}
%\vspace{-40pt}
\vspace{-5pt}

By optimizing multi-round denoising process, SRRL gains self-reflection ability through PRMs. However, a problem is how to connect the multiple rounds of denoising processes, allowing reflective iteration between image CoT steps.
%Diffusion models suffer from the issue that denoising process is a single round procedure, and reward evaluation can only be performed on the final image, limiting its self-reflection capability. 
To achieve multi-round self-reflection between different denoising trajectories, SRRL proposes condition guided forward process, which adds conditional noise to intermediate samples at the end of each denoising round to obtain noisy samples for the next round of reflective denoising process.
% Diffusion denoising process cannot iterate once denoising is complete, limiting its self-reflection capability. To achieve multi-round self-reflection, condition guided forward process performs a forward process on the intermediate samples at the end of each denoising round to obtain the initial noisy samples for the next round of reflection. CFG~\cite{ho2022classifier} provides a way to inject text conditions, and the position of initial noisy samples carries text conditions and the guidance gap between denoising and forward process captures semantic information~\cite{bai2024zigzag}. Inspired by the phenomenon, SRRL injects text condition during the forward process through the guidance gap.

Given the intermediate sample $x_0^k$, the condition guided forward process aims to add noise to obtain the noisy sample $x_T^{k+1}$ of the next round, which can be formulated as:

\vspace{-10pt}
\begin{equation}
    \begin{aligned}
        x_T^{k+1} &= \prod_{t=1}^T \chi(\tilde{x}_t^k|\tilde{x}_{t-1}^k,c), \quad k=0,1,\cdots, K \\ 
        \chi(\tilde{x}_t^k|\tilde{x}_{t-1}^k,c) &= \sqrt{\frac{\alpha_t}{\bar{\alpha}_{t-1}}}\tilde{x}_{t-1}^k + (\frac{1-\alpha_t}{\sqrt{1-\bar{\alpha}}_t}- \sqrt{\frac{\alpha_t (1-\bar{\alpha}_{t-1})}{\bar{\alpha}_{t-1}}})\tilde{\epsilon}_\theta(\tilde{x}_{t-1}^k,c,t,\lambda) \\
        %\chi(\tilde{x}_t^k|\tilde{x}_{t-1}^k,c) &=  \sqrt{\frac{\alpha_t}{\alpha_{t-1}}} \tilde{x}_{t-1}^k +\sqrt{\alpha_t} \big(  \sqrt{\frac{1}{\alpha_t}-1} -\sqrt{\frac{1}{\alpha_{t-1}}-1} \big)\tilde\epsilon_\theta( \tilde{x}_{t-1}^k,c,t,\lambda)
    \end{aligned}
\label{equ:add_noise}
\end{equation}

% \resizebox{1\hsize}{!}{$
%     x_T^{k+1}= \prod_{t=1}^T \chi(\tilde{x}_t^k|\tilde{x}_{t-1}^k,c), \quad \quad \chi(\tilde{x}_t^k|\tilde{x}_{t-1}^k,c)=  \sqrt{\frac{\alpha_t}{\alpha_{t-1}}} \tilde{x}_{t-1}^k +\sqrt{\alpha_t} \left(  \sqrt{\frac{1}{\alpha_t}-1} -\sqrt{\frac{1}{\alpha_{t-1}}-1} \right)\tilde\epsilon_\theta( \tilde{x}_{t-1}^k,c,t,\lambda)
% $}
% %\end{equation}
where $x_T^{k+1}=\tilde{x}_T^k$ and $x_0^k= \tilde{x}_0^k$. SRRL sets CFG guidance scale $\lambda$ in forward process smaller than that in denoising process, e.g. 1.0 and 4.5. By creating a guidance gap between forward process and denoising process, SRRL injects text condition during forward process, leading to progressively better results with more reflection rounds.
We use denoising diffusion implicit model (DDIM)~\cite{song2020denoising} inversion scheduler, which is a deterministic sampling method to precisely inject text conditions.

%%% What COT mean
In summary, SRRL optimizes the denoising trajectory over multiple rounds and introduces intermediate sample reward evaluations, which addresses the issue that reward prediction is limited to final images. Besides, by introducing condition guided forward process, SRRL establishes inter-trajectory CoT connections, enabling iterative reflection and knowledge transfer across sequential steps. 
Multiple rounds of the denoising and forward process provide self-reflection ability, facilitating image reasoning generation in diffusion models. The pseudo-codes of training and inference process of SRRL are shown in Algorithm~\ref{alg:train} and Algorithm~\ref{alg:inference}.

\vspace{-4pt}
\section{Experiments} \label{sec:experiments}
\vspace{-4pt}

In this section, we evaluate SRRL's effectiveness in image reasoning generation tasks. We aim to answer the following questions: 
i) Is it possible to leverage a self-reflective reinforcement learning algorithm to achieve image reasoning generation adhered to physical laws and unconventional physical phenomena?
ii) How do generated images include reasoning and thought processes?
%iii) Whether different reward evaluation models affect the learning of model reasoning abilities?

\begin{figure}
    \centering
    \includegraphics[width=1\textwidth]{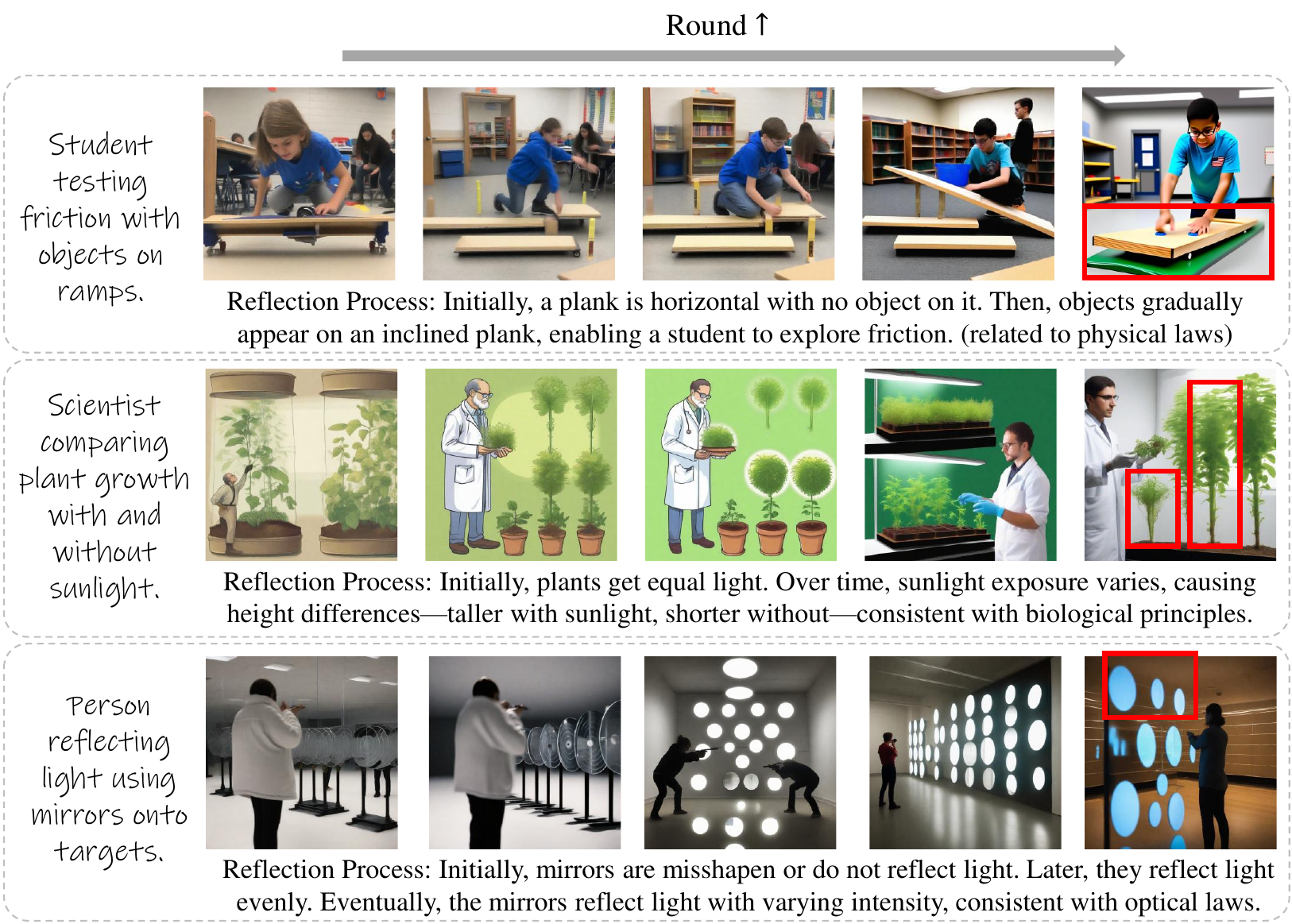}
    \vspace{-16pt}
    \caption{
     Reasoning generation of images related to physical laws.
    } 
    \label{fig:physic}
\vspace{-16pt}
\end{figure}

\vspace{-4pt}
\subsection{Experimental Setup}
\vspace{-3pt}
    
\textbf{T2I diffusion models}. We use Stable Diffusion (SD) v1.4~\cite{rombach2022high} and SD XL~\cite{podell2023sdxl} as the backbone diffusion models, which are open-source and widely used for T2I tasks. We perform LoRA~\cite{hu2022lora} fine-tuning on U-Net in SD, which is a method that saves GPU memory and accelerates training efficiency. During multi-round reflective denoising process, SRRL uses DDIM~\cite{song2020denoising} scheduler. During condition guided forward process, SRRL uses DDIM inversion scheduler. The number of sampling steps is set to 20. The implementation details are shown in Appendix~\ref{appen:imple}.

\textbf{Reward models and metrics}. We use CLIP Score~\cite{hessel2021clipscore}, ImageReward~\cite{xu2023imagereward}, and VQAScore~\cite{lin2024evaluating} to evaluate the text-image alignment and image reasoning abilities of models. CLIP Score measures the similarity between text and image embeddings via CLIP model~\cite{radford2021learning}, trained with contrastive learning for cross-modal alignment. 
%We use ViT-H-14 CLIP model. 
Image reward~\cite{xu2023imagereward} evaluates the
general-purpose text-to-image human preference by training on total 137k pairs text-images with expert comparisons. 
VQAScore~\cite{lin2024evaluating} employs a visual-question-answering model to compute an alignment score. This is achieved by measuring the probability of the model responding 'Yes' to the question: 'Does this figure depict \{text\}?'.    
VQAScore is better in evaluating image reasoning ability due to its judgment ability.

\textbf{Prompt type}. We evaluate the effectiveness of our algorithm on three types of prompts. 
i) Following previous works~\cite{black2023training,hu2025towards}, we use the prompt template "a(n) [animal] [activity]", which evaluates text-image alignment. There are 45 kinds of animals and three activities: “riding a bike”, “playing chess”, and “washing dishes”. Animals and activities are randomly matched.
ii) Physical phenomenon-related prompts. These prompts include knowledge about physical laws. Details are shown in Appendix~\ref{appen:prompt}.
iii) Unconventional physical phenomena prompts. These prompts contradict common phenomena to evaluate models' imagination capabilities. Details are shown in Appendix~\ref{appen:prompt}.
It is worth noting that image reasoning capability and image-text alignment are not equivalent, and we discuss it in Sec.~\ref{sec:discussion}.

%\textbf{Metrics}. We evaluate the generated images using CLIP Score~\cite{hessel2021clipscore}, ImageReward~\cite{xu2023imagereward}, and VQAScore~\cite{lin2024evaluating}. 
%CLIP Score evaluates text-image alignment, ImageReward evaluates human preference, and VQAScore evaluates visual-question-answering. VQAScore is better in evaluating image reasoning ability due to its judgment ability.

%A higher CLIP score, higher image reward, and higher VQAScore indicate better image quality (text-image alignment and alignment with human feedback).

\vspace{-4pt}
\subsection{Physical Law Related Image Generation}
\vspace{-3pt}
% \begin{figure}
%     \centering
%     \includegraphics[width=1\textwidth]{images/physic_figure_2.pdf}
%     \caption{
%      Generated images related to physical laws.
%     } 
%     \label{fig:physic}
% \vspace{-4pt}
% \end{figure}

%We fine-tune SD XL~\cite{podell2023sdxl} using SRRL algorithm on physical phenomenon related prompts. Some qualitative results are shown in Fig.~\ref{fig:physic}. 
We train SD XL \cite{podell2023sdxl} with the SRRL algorithm using prompts related to physical phenomena. Fig.~\ref{fig:physic} shows some qualitative results.
The first prompt is "Student testing friction with objects on ramps". Initially, generated images lack inclined planks, and objects on the plank are unclear. With iterative self-reflection training, the final image includes an inclined plank with clear objects on it, depicting a student testing friction.
The second prompt is "Scientist comparing plant growth with and without sunlight", which contains biological principles: plants receiving adequate sunlight grow better than those that do not. At first, two plants are similar. Gradually, the model learns to differentiate the intensity of light exposure on plants. Eventually, the model realizes that plants exposed to more light grow better.
The third prompt is "Person reflecting light using mirrors onto targets". It indicates that the light on different mirrors is different. Initially, the mirrors are misshapen or do not reflect any light. Later, they reflect light evenly. Eventually, the mirrors reflect light with varying intensity, consistent with physical laws.
The above results indicate that SRRL, through self-reflection, can gradually learn to reason and generate images following physical laws.

\vspace{-4pt}
\subsection{Unconventional Image Generation}
\vspace{-3pt}

\begin{figure}
    \centering
    \includegraphics[width=1\textwidth]{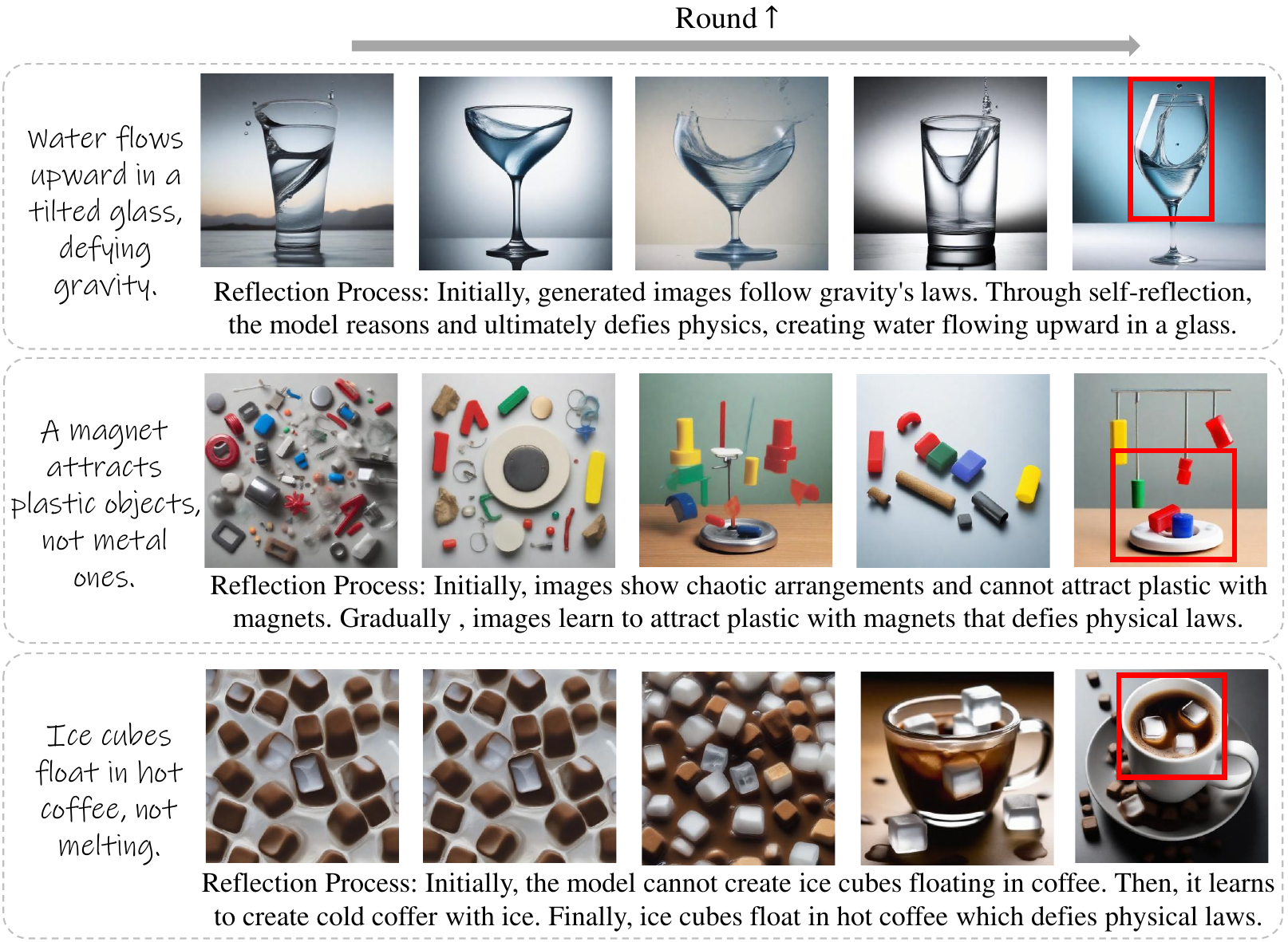}
    \vspace{-16pt}
    \caption{
     Reasoning generation of images related to unconventional physical phenomena.
    } 
    \label{fig:unconventional}
\vspace{-16pt}
\end{figure}
We train SD XL model~\cite{podell2023sdxl} with SRRL algorithm using prompts for unconventional physical phenomena, which are counterintuitive and contradict usual physical phenomena. Some qualitative results are shown in Fig.~\ref{fig:unconventional}.  
The first prompt is "Water flows upward in a tilted glass, defying gravity". At first, generated images obey physical laws of gravity. As the self-reflection process continues, the model engages in reasoning and eventually overcomes physical laws, generating an image of water flowing upwards in a glass. 
The second prompt is "A magnet attracts plastic objects, not metal ones". Initially, the objects in the images are chaotic, indicating that the model does not know how to use a magnet to attract plastic objects. Through self-reflection process, the model learns to attract plastic objects with a magnet, even though this defies physical laws.
The third prompt is "Ice cubes float in hot coffee, not melting". The model initially cannot generate ice cubes floating in hot coffee. As the reasoning process progresses, the model learns this concept. From the bubbling coffee in the image, it can be inferred that the coffee is hot, while the ice cubes in the coffee have not melted. 
From the results above, it can be observed that initially, the model either adheres to physical laws or lacks relevant knowledge. As self-reflection activates its reasoning abilities, the model is able to generate images that defy common sense or physical laws.

\vspace{-4pt}
\subsection{Visualization of Image Reasoning Process}
\vspace{-3pt}

\begin{figure}
    \centering
    \includegraphics[width=1\textwidth]{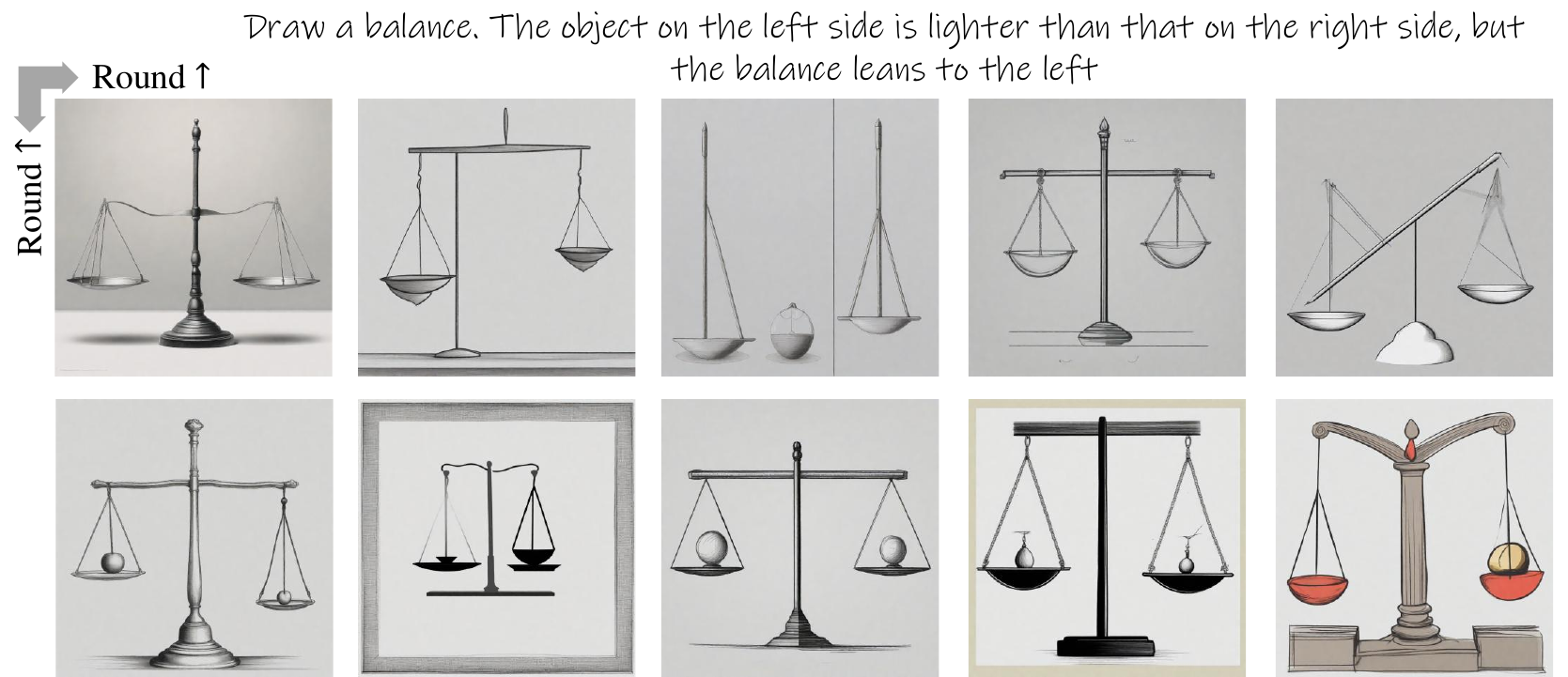}
    \vspace{-16pt}
    \caption{
     Reasoning generation process of the prompt related to a balance. Initially, the model generates an image of a balance tilted left without objects or tilted right with lighter objects on the left and heavier ones on the right, both following physical laws. Eventually, it learns to create images defying logic: a balance tilts left with no objects on the left and a small ball on the right.
    } 
    \label{fig:reason_process}
\vspace{-14pt}
\end{figure}

\begin{figure}
    \centering
    \includegraphics[width=1\textwidth]{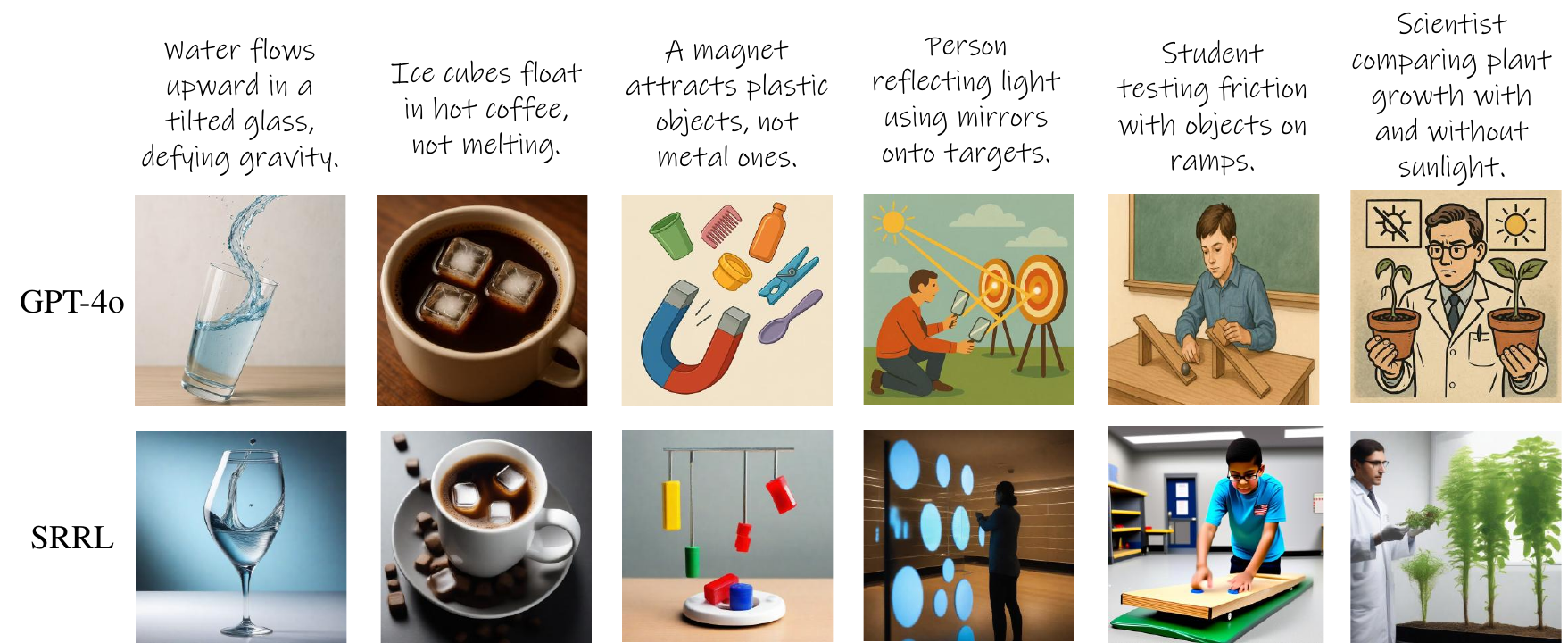}
    \vspace{-16pt}
    \caption{
    Example of generated images of physical phenomenon related prompts and unconventional physical phenomena prompts by GPT-4o~\cite{hurst2024gpt} and SRRL. 
    } 
    \label{fig:gpt4o}
\vspace{-18pt}
\end{figure}

To visualize the reasoning process of SRRL, we incorporate two prompts: "Draw a balance. The object on the left side is lighter than that on the right side, but the balance leans to the left" and "Cars are driving on water", which contains contradictory knowledge. Reasoning process visualizations of the prompt related to the balance is shown in Fig.~\ref{fig:reason_process}. We train SD XL model~\cite{podell2023sdxl} with SRRL algorithm on one prompt each time and we use ImageReward~\cite{xu2023imagereward} as the reward model. 
For results of the first prompt in Fig.~\ref{fig:reason_process}, 
%initially, the model either generates an image of a balance tilted to the left with no objects on it, or generates an image of a balance tilted to the right with objects that are lighter on the left and heavier on the right. Both types of images conform to common sense or physical laws. Finally, the model learns to generate images containing contradictory knowledge: the balance is tilted to the left although there are no objects on the left side and a small ball on the right side. 
initially, the model generates images of a balance either tilted left with no objects or tilted right with lighter objects on the left and heavier ones on the right, both aligning with common sense or physical laws. Eventually, the model produces an image with contradictory elements: the balance is tilted left despite having no objects on the left and a small ball on the right.
More reasoning process visualization and analysis are shown in Appendix~\ref{appen:reasoning_process}.
%For results of the second prompt in Fig.~\ref{fig:reason_process_2}. The general common sense is that cars drive on land, but the prompt requires generating an image of a car driving on water. In the initially generated images, the car appears to fly out of the water. In the final generated images, the car drives in the center of the lake rather than floats. 
This suggests that by introducing self-reflection mechanism, the model can perform reasoning and has the ability to generate images adhering to contradictory common sense, showing the model's imagination ability.

\vspace{-4pt}
\subsection{Comparison with Baselines}
\vspace{-4pt}
% \begin{figure}
%     \centering
%     \includegraphics[width=1\textwidth]{images/gpt4o.pdf}
%     \caption{
%     Example of generated images of physical phenomenon related prompts and unconventional physical phenomena prompts by GPT-4o~\cite{hurst2024gpt} and SRRL. 
%     } 
%     \label{fig:gpt4o}
% \vspace{-4pt}
% \end{figure}

We compare samples generated by SRRL and GPT-4o~\cite{hurst2024gpt}, which is the most advanced image generation model recently. Results are shown in Fig.~\ref{fig:gpt4o}. SRRL generates similar or higher quality images compared to GPT-4o, showing reasoning capabilities akin to those of GPT-4o. Furthermore, while GPT-4o generates images in a cartoon style, SRRL reasons high-quality realistic images. 

\begin{figure}[t]
    \centering
    \includegraphics[width=1\textwidth]{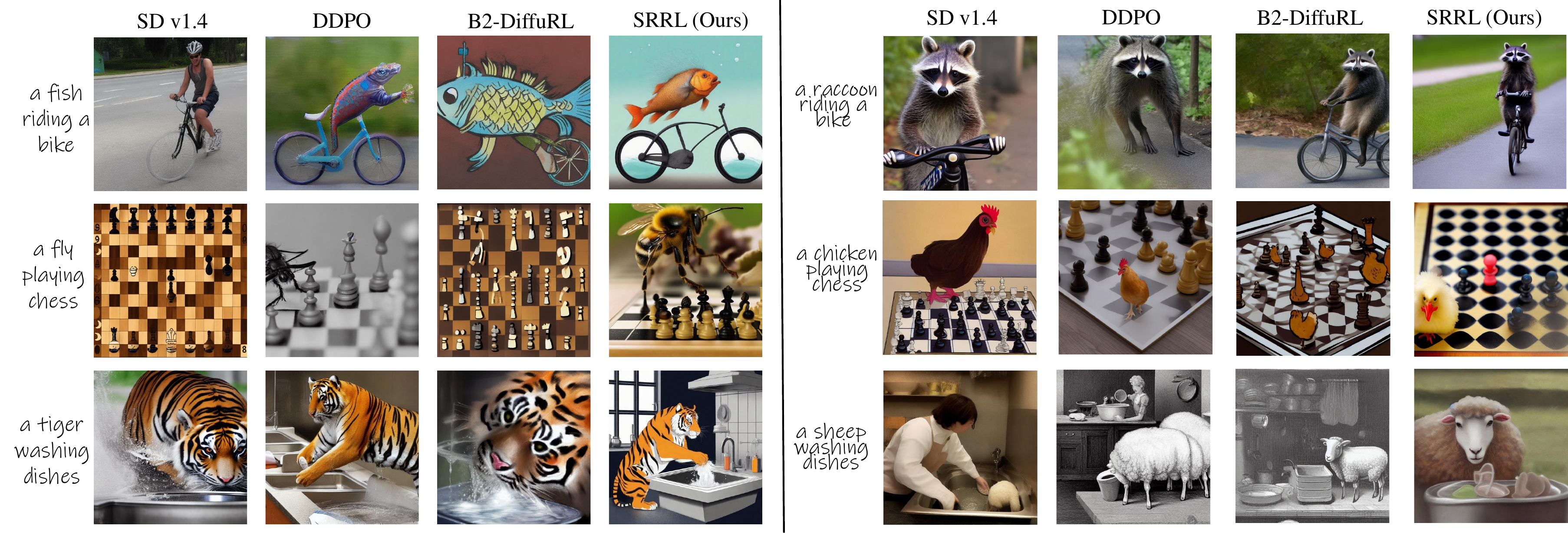}
    \vspace{-16pt}
    \caption{
    Examples of prompt template "a(n) [animal] [activity]" by baselines~\cite{rombach2022high,black2023training,hu2025towards} and SRRL. 
    } 
    \label{fig:compare_baseline}
\vspace{-7pt}
\end{figure}

We train SD v1.4 model using SRRL algorithm on the prompt template "a(n) [animal] [activity]" to compare with previous works~\cite{black2023training,hu2025towards}. Fig.~\ref{fig:compare_baseline} shows some results of the prompt template.
Compared to the baselines, the images generated by SRRL are better aligned with prompts and are of higher quality. This indicates that after introducing self-reflection mechanism, the model's ability of traditional text-to-image alignment also improves. 

% \usepackage{booktabs}
% \usepackage{booktabs}
\begin{table}[t]
\centering
\resizebox{0.65\hsize}{!}{
\begin{tabular}{@{}c|cccc@{}}
\toprule
Methods     & SD & DDPO~\cite{black2023training}   & B2-DiffuRL~\cite{hu2025towards} & SRRL (Ours) \\ \midrule
CLIP Score $\uparrow$  & 0.3624  & \textbf{0.3683} & 0.3674     & 0.3662      \\
ImageReward $\uparrow$ & 0.2823  & 0.3534 & 0.3682     & \textbf{0.3807}      \\
VQAScore  $\uparrow$   & 0.6045  & 0.6145 & 0.6174     & \textbf{0.6338}      \\ \midrule
\end{tabular}
}
\caption{Quantitative results of prompt template "a(n) [animal] [activity]" on different metrics. All experiments are done based on SD v1.4.}
\vspace{-16pt}
\label{tab: sd14}
\end{table}

% \begin{figure}[!htbp]
%     \centering
%     \includegraphics[width=1\textwidth]{images/output_table.pdf}
%     \vspace{-18pt}
%     \caption{
%     Performance of multi-round self-reflection of SRRL. The left is the results of physical phenomenon related prompts, and the right is those of unconventional physical phenomena prompts. All experiments are done based on SD XL.
%     } 
%     \label{fig:output_table}
% \vspace{-16pt}
% \end{figure}

% \begin{figure}[!htbp]
%     \centering
%     \begin{minipage}{0.57\textwidth}
%         \includegraphics[width=\linewidth]{images/output_table.pdf}
%         %\caption*{(a) Physical phenomenon related prompts}
%     \end{minipage}
%     \hfill
%     \rule[-1.1cm]{0.4pt}{2.8cm}
%     \raisebox{3mm}{
%     \begin{minipage}{0.39\textwidth}
%         \includegraphics[width=\linewidth]{images/output_table_2.pdf}
%        %     \caption*{(b) Unconventional physical phenomena prompts}
%     \end{minipage}
%     }
%     \vspace{-5pt}
%     \caption{
%     Performance of multi-round self-reflection of SRRL. Left. Performance of multi-round self-reflection of SRRL. The left is the results of physical phenomenon related prompts, and the right is those of unconventional physical phenomena prompts. 
%     Right. The top is a sample of physical phenomenon related prompts, which reflects magnet images. The bottom is a sample of unconventional physical phenomena prompts, which reflect a book remains dry underwater.    
%     All experiments are done based on SD XL.  
%     } 
%     \label{fig:output_table}
%     \vspace{-16pt}
% \end{figure}

\begin{figure}[!htbp]
    \centering
    \hspace{0.055\textwidth}
    \begin{subfigure}{0.4\textwidth}
        \centering
        \includegraphics[width=\textwidth]{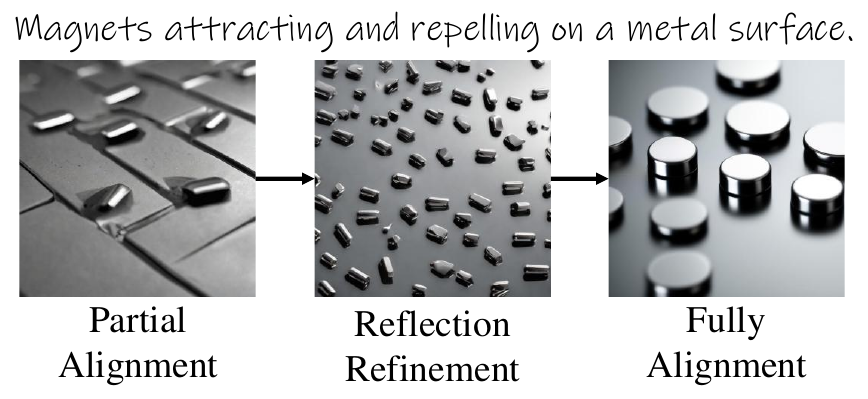} % 替换为你的图片文件
        %\caption{左边的子图}
        %\label{fig:sub1}
    \end{subfigure}
    %\hfill
    \hspace{0.105\textwidth}
    \begin{subfigure}{0.4\textwidth}
        \centering
        \includegraphics[width=\textwidth]{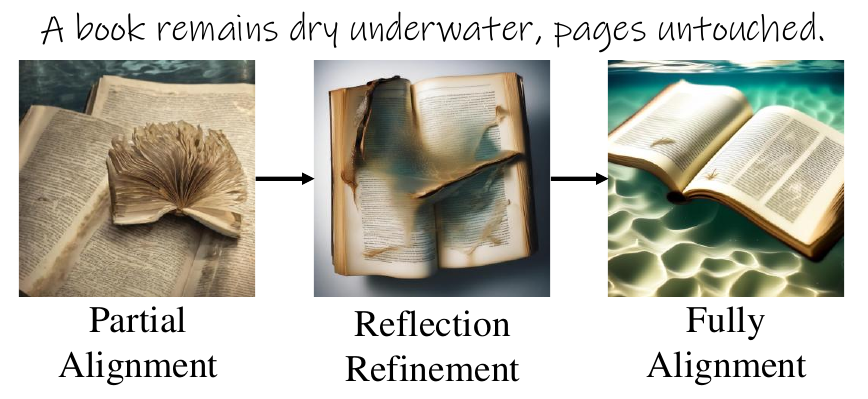}
        %\caption{右边的子图}
        %\label{fig:sub2}
    \end{subfigure}

    \vspace{-1.23mm} % 根据需要调整垂直间距

    \begin{minipage}{\textwidth}
        \centering
        \includegraphics[width=1\textwidth]{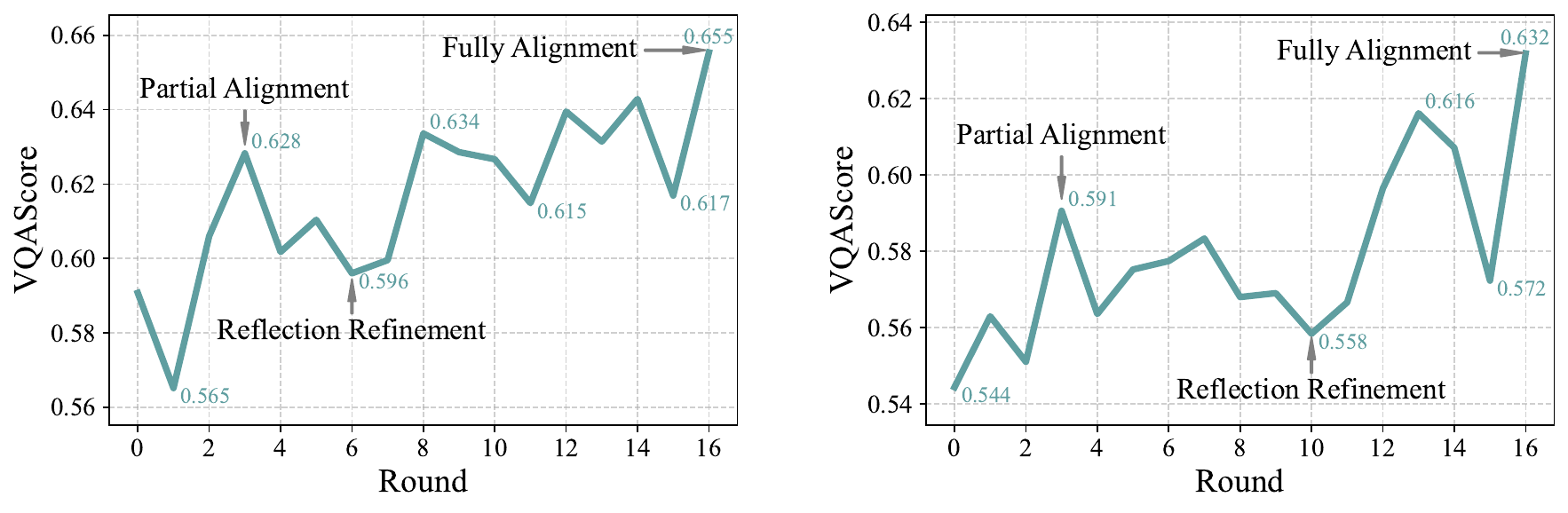} 
        %\caption{下面的整图}
        %\label{fig:sub3}
    \end{minipage}
    \vspace{-11pt}
    \caption{Performance of multi-round self-reflection of SRRL. The left is results of physical phenomenon related prompts, and the right is those of unconventional physical phenomena prompts. The figure above shows some cases of reflection process. All experiments are done based on SD XL.}
    \label{fig:output_table}
\vspace{-16pt}
\end{figure}

\vspace{-4pt}
\subsection{Ablation of Reward Models and Self-Reflection Rounds}
\vspace{-3pt}
% \begin{figure}
%     \centering
%     \includegraphics[width=1\textwidth]{images/output_table.pdf}
%     \caption{
%     Performance of multi-round self-reflection of SRRL. The left is the results of physical phenomenon related prompts, and the right is those of unconventional physical phenomena prompts. All experiments are done on SD XL.
%     } 
%     \label{fig:output_table}
% \vspace{-4pt}
% \end{figure}

SRRL uses CLIP Score, ImageReward, and VQAScore as reward models, and we compare the impact of different reward models on the model's learning and reasoning ability. We find that ImageReward and VQAScore perform better in enhancing the model's generation quality, while CLIP Score tends to degrade when the number of training epochs is too high. Quantitative results of prompt template  "a(n) [animal] [activity]" are shown in Tab.~\ref{tab: sd14}. Compared with baseline, SRRL performs better in ImageReward and VQAScore. 

We evaluate the performance of SRRL on physical phenomenon related prompts and unconventional physical phenomena prompts, and the results are shown in Fig.~\ref{fig:output_table}. The quality of generated images improves as the round increases with self-reflection process. We also notice the phenomenon of reflection refinement, which involves adjusting the generated images through image reconstruction.
%The experiment is done on SD XL.

%\subsection{Qualitative Results}

%\subsection{Quantitative Results}

%\vspace{-100pt}
\vspace{-6pt}
\section{Discussion and Conclusion}
\vspace{-4pt}
\label{sec:discussion}

% Recently, chain-of-thought (CoT) has been widely applied in the field of large language models, significantly enhancing self-reflective abilities in complex tasks. If CoT is introduced into the image generation field, a question is which specific image generation problems CoT should be used to address. In this article, we explore generating images related to physical laws. Compared to other types of images, these images better reflect the model's reasoning and imagination abilities. Analogous to text generation, we believe that generating images related to physical laws could be a worthy challenge for image generation when introducing CoT, beyond merely improving the degree of text-image alignment. 
% Exploring application scenarios for CoT in image generation, such as creating complex logical images, is an interesting problem which can be explored.

Recently, chain-of-thought (CoT) has been widely adopted in large language models, enhancing their self-reflective abilities in complex tasks. When applied to image generation, a key question is which challenges CoT should address. This paper explores using CoT to generate images that adhere to physical laws, as these images better showcase models' reasoning and imagination. Similar to complex NLP problems, creating images adhered to physical laws presents a challenge beyond only improving text-image alignment
because physical laws are usually implicit in textual descriptions.
Exploring CoT's potential in generating logical images is an intriguing task for further research.

%This paper introduces SRRL, a novel self-reflective reinforcement learning algorithm for diffusion models. Our goal is to enhance reasoning in these models using image CoT and self-reflection mechanisms. SRRL employs multi-round reflective denoising process and condition-guided forward process, enabling reasoning through iterative self-reflection. Experiments demonstrate that SRRL-trained models can produce images aligned with physical laws and unconventional scenarios, demonstrating the image reasoning ability.
This paper presents SRRL, a novel self-reflective reinforcement learning algorithm for diffusion models aimed at improving reasoning abilities. By introducing image CoT and self-reflection mechanisms, SRRL proposes multi-round reflective denoising process and condition guided forward process. Experiments show that SRRL-trained models generate images that adhere to physical laws and unconventional scenarios, showcasing image reasoning abilities.

\begin{ack}
This work is supported by the National Science and Technology Major Project (2023ZD0121403), National Natural Science Foundation of China (No. 62406161), China Postdoctoral Science Foundation (No. 2023M741950), and the Postdoctoral Fellowship Program of CPSF (No. GZB20230347).
\end{ack}
\bibliographystyle{plain}
\bibliography{mm_ref}

%%%%%%%%%%%%%%%%%%%%%%%%%%%%%%%%%%%%%%%%%%%%%%%%%%%%%%%%%%%%

\appendix

\section{Derivations} \label{appen:derivations}

\subsection{Equation \ref{equ:policy_gradient}}

\resizebox{1\hsize}{!}{$
    \begin{aligned}
        \nabla_\theta [-\mathcal{J}_{SRRL}(\theta)] &= \nabla_\theta \mathbb{E}_{p(c)} \mathbb{E}_{p_\theta(x_0|c)}\mathbb{E}_{k\sim U(0,K)}[-r(x_0^k,c)] \\
         &= -\mathbb{E}_{p(c)}\mathbb{E}_{k\sim U(0,K)} [\nabla_\theta \int r(x_0^k,c)p_\theta (x_0^k|c)dx_0^k] \\ 
         &= -\mathbb{E}_{p(c)} \mathbb{E}_{k\sim U(0,K)} [\nabla_\theta \int r(x_0^k,c)(\int p_\theta(x_{0:T}^k|c)dx_{1:T}^k)dx_0^k] \\
         &= -\mathbb{E}_{p(c)}\mathbb{E}_{k\sim U(0,K)} [\int \nabla_\theta \log p_\theta (x_{0:T}^k|c) r(x_0^k,c) p_\theta(x_{0:T}^k|c)dx_{0:T}^k] \\
         &= -\mathbb{E}_{p(c)}\mathbb{E}_{k\sim U(0,K)} [\int \nabla_\theta \log \left(p_T (x_{T}^k|c) \prod_{t=1}^T p_\theta (x_{t-1}^k|x_t^k,c)\right) r(x_0^k,c) p_\theta(x_{0:T}^k|c)dx_{0:T}^k] \\ 
         &= -\mathbb{E}_{p(c)}\mathbb{E}_{p_\theta(x_{0:T}^k|c)}\mathbb{E}_{k\sim U(0,K)}[\sum_{t=1}^T \nabla_\theta \log p_\theta(x_{t-1}^k|x_t^k,c)r(x_0^k,c)]
    \end{aligned}
$}

Here the proof uses the continuous assumptions of $p_\theta(x_{0:T}^k|c)r(x_0^k,c)$.

\subsection{ Equation \ref{equ:add_noise}}

Following DDPM~\cite{ho2020denoising}, the denoising process is formulated as:

\begin{equation}
    x_{t-1}^k= \frac{\sqrt{\bar{\alpha}_{t-1}}}{\sqrt{\alpha_t}}(x_t^k- \frac{1-\alpha_t}{\sqrt{1-\bar{\alpha}_t}}\epsilon_\theta(x_t^k,c,t,\lambda))+\sqrt{1-\bar{\alpha}_{t-1}}\epsilon_\theta(x_t^k,c,t,\lambda)
\end{equation}

Then, solve for $x_{t}$ based on $x_{t-1}$,

\begin{equation}
    \tilde{x}_t^k= \sqrt{\frac{\alpha_t}{\bar{\alpha}_{t-1}}}\tilde{x}_{t-1}^k + (\frac{1-\alpha_t}{\sqrt{1-\bar{\alpha}}_t}- \sqrt{\frac{\alpha_t (1-\bar{\alpha}_{t-1})}{\bar{\alpha}_{t-1}}})\tilde{\epsilon}_\theta(\tilde{x}_{t-1}^k,c,t,\lambda)
\end{equation}

Here we leverage the assumption that $\tilde{\epsilon}_\theta(\tilde{x}_{t-1}^k,c,t,\lambda) \approx \tilde{\epsilon}_\theta(\tilde{x}_{t}^k,c,t,\lambda)$.

We can inject condition if there is a guidance gap between forward process and denoising process.

For convenience, we set:
\begin{equation}
    \gamma_t=\sqrt{\frac{\alpha_t}{\bar{\alpha}_{t-1}}}, \quad \quad \eta_t= (\frac{1-\alpha_t}{\sqrt{1-\bar{\alpha}}_t}- \sqrt{\frac{\alpha_t (1-\bar{\alpha}_{t-1})}{\bar{\alpha}_{t-1}}})
\end{equation}

Then,

\begin{equation}
    \begin{aligned}
        \tilde{x}_T^k &= \gamma_T \tilde{x}_{T-1}^k+ \eta_T \tilde\epsilon_\theta(\tilde{x}_{T-1}^k,c,t,\lambda) \\
        &= \gamma_T\gamma_{T-1} \tilde{x}_{T-2}^k+  \gamma_T \eta_{T-1} \tilde\epsilon_\theta(\tilde{x}_{T-2}^k,c,t,\lambda) +   \eta_T \tilde\epsilon_\theta(\tilde{x}_{T-1}^k,c,t,\lambda) \\
        &= \cdots \\
        &= \prod_{i=0}^T \gamma_i \tilde{x}_0 +\sum_{t=1}^T \eta_t \prod_{l=t+1}^T \gamma_l \epsilon_\theta(\tilde{x}_{t-1}^k,c,t,\lambda)
        \end{aligned}
\end{equation}

Similarly, we can get:
\begin{equation}
    x_T^k= \prod_{i=0}^T \gamma_i x_0 +\sum_{t=1}^T \eta_t \prod_{k=t+1}^T \eta_k \epsilon_\theta(x_{t-1}^k,c,t,\lambda_{\text{Forward}})
\end{equation}

The guidance gap can be formulated as:
\begin{equation}
    \begin{aligned}
        \delta_k &= (x_T^k- \tilde{x}_T^k)^2 \\
        &= [ \prod_{i=0}^T \gamma_i (x_0-\tilde{x}_0) +\sum_{t=1}^T \eta_t \prod_{l=t+1}^T \gamma_l (\epsilon_\theta(x_{t-1}^k,c,t,\lambda_{\text{Forward}})-\epsilon_\theta(\tilde{x}_{t-1}^k,c,t,\lambda)) ]^2 \\
        &= (\sum_{t=1}^T F(\eta_t,\gamma_t)(\epsilon_\theta(x_{t-1}^k,c,t,\lambda_{\text{Forward}})-\epsilon_\theta(\tilde{x}_{t-1}^k,c,t,\lambda)))^2 \\
        &= (\sum_{t=1}^T F(\eta_t,\gamma_t)(\lambda_{\text{Forward}}-\lambda)\epsilon_\theta(x_{t-1}^k,c,t))^2
    \end{aligned}
\end{equation}

By setting a guidance scale gap between $\lambda$ and $\lambda_{\text{Forward}}$, we can inject text condition during condition injection reflection forward process. Through multiple rounds of self reflection, the effect of condition injection is enhanced.

\section{Implementation Details}\label{appen:imple}

Our experiments are all done on NVIDIA RTX 4090 24G GPUs. Each round of the training process takes approximately 20 minutes.

When fine-tuning Stable Diffusion model~\cite{rombach2022high,podell2023sdxl} using LoRA according to SRRL algorithm, the configs are:

\begin{table}[!htbp]
\centering  
\begin{tabular}{@{}c|c@{}}
\toprule
Config         & Value                                     \\ \midrule
LoRA rank      &   4  \\
LoRA alpha     &   4   \\
lr             & 1e-4 \\
optimizer     &  Adam~\cite{kingma2014adam}                                         \\
weight decay of optimizer      & 1e-4                                        \\
$\beta_1,\beta_2$        & (0.9,0.999)                                       \\

number of samples per batch $G$       & 32                                  \\
self-reflection total rounds $K$      & 10                                      \\
denoising timestep $T$     & 20                                         \\
reward function $r$  & CLIP Score~\cite{hessel2021clipscore}, ImageReward~\cite{xu2023imagereward}, VQAScore~\cite{lin2024evaluating}                                     \\
training epoch number $E$  & 2                                    \\
forward guidance scale &  0.5  \\
denoising guidance scale  & 3.0 \\
inference guidance scale & 7.5 \\
\bottomrule
\end{tabular}
\end{table}

\section{Pseudo-code of SRRL}\label{appen:pseudocode}

\begin{algorithm}[H]
\caption{SRRL Training Process}
\label{alg:train}
\begin{algorithmic}[1]
\InCom Pretrained diffusion model $p_\theta$, denoising timestep $T$, self-reflection total rounds $K$,  number of samples per batch $G$, prompts list C, reward function $r$, training epoch number $E$. 

\STATE $k=0$
\REPEAT 
\STATE $e=0$
\REPEAT
\STATE SampleList=[]
\STATE $n=0$
\REPEAT 
\STATE Random choose prompt c from C,
\STATE Random sample Gaussian noise $x_T$ in $\mathcal{N}(0,1)$,
\STATE $i=0$
\REPEAT
\STATE Denoise $x_T^i$ to $x_0^i$ with $p_\theta$,
\STATE Noise injection $x_0^i$ to $x_T^{i+1}$ with $p_\theta$,
\STATE $i=i+1$
\UNTIL $i=k$
\STATE $x_{T:0}^k= \text{DDIM-Scheduler}_\theta(x_T^k \rightarrow x_0^k)$,
\STATE SampleList.append([$x_{T:0}^k$,c]),
\STATE $n=n+1$
\UNTIL $n=G$
\STATE Evaluate $r(x_{0,i=1:G}^k,c)$,
\STATE $\text{score}_{i=1}^G= \text{Reward Normalization}(r(x_{0,i=1:G}^k,c)$,
\STATE $\text{score}_{\max},i_{\max}= \text{maximum}(\text{score}_{i=1}^G),\text{index}(\text{score}_{\max})$,
\STATE $\text{score}_{\min},i_{\min}= \text{minimum}(\text{score}_{i=1}^G),\text{index}(\text{score}_{\min})$,
\STATE update $\theta$ according to $[x_{T:0, i_{\max}}^k, \text{score}_{\max},x_{T:0, i_{\min}}^k, \text{score}_{\min},c]$.

\STATE $e=e+1$
\UNTIL $e=E$
\STATE $k=k+1$
\UNTIL $k=K$
\OutCom Fine-tuned Model $p_{\theta'}$

\end{algorithmic}
%\label{alg:train}
\end{algorithm}

\begin{algorithm}[H]
\caption{SRRL Inference Process}
\label{alg:inference}
\begin{algorithmic}[1]
\InCom Fine-tuned diffusion model $p_\theta$, self-reflection rounds $k$
\STATE Random sample Gaussian noise $x_T^0$ in $\mathcal{N}(0,1)$,
\STATE $i=0$
\REPEAT
\STATE Denoise $x_T^i$ to $x_0^i$ with $p_\theta$,
\STATE Noise injection $x_0^i$ to $x_T^{i+1}$ with $p_\theta$,
\STATE $i=i+1$
\UNTIL $i=k$
\STATE Denoise $x_T^k$ to $x_0^k$
\OutCom Self-reflective images $x_0^k$

\end{algorithmic}
%\label{alg:inference}
\end{algorithm}

\section{Additional Visualization of Image Reasoning Process}\label{appen:reasoning_process}

\begin{figure}[!htbp]
    \centering
    \includegraphics[width=1\textwidth]{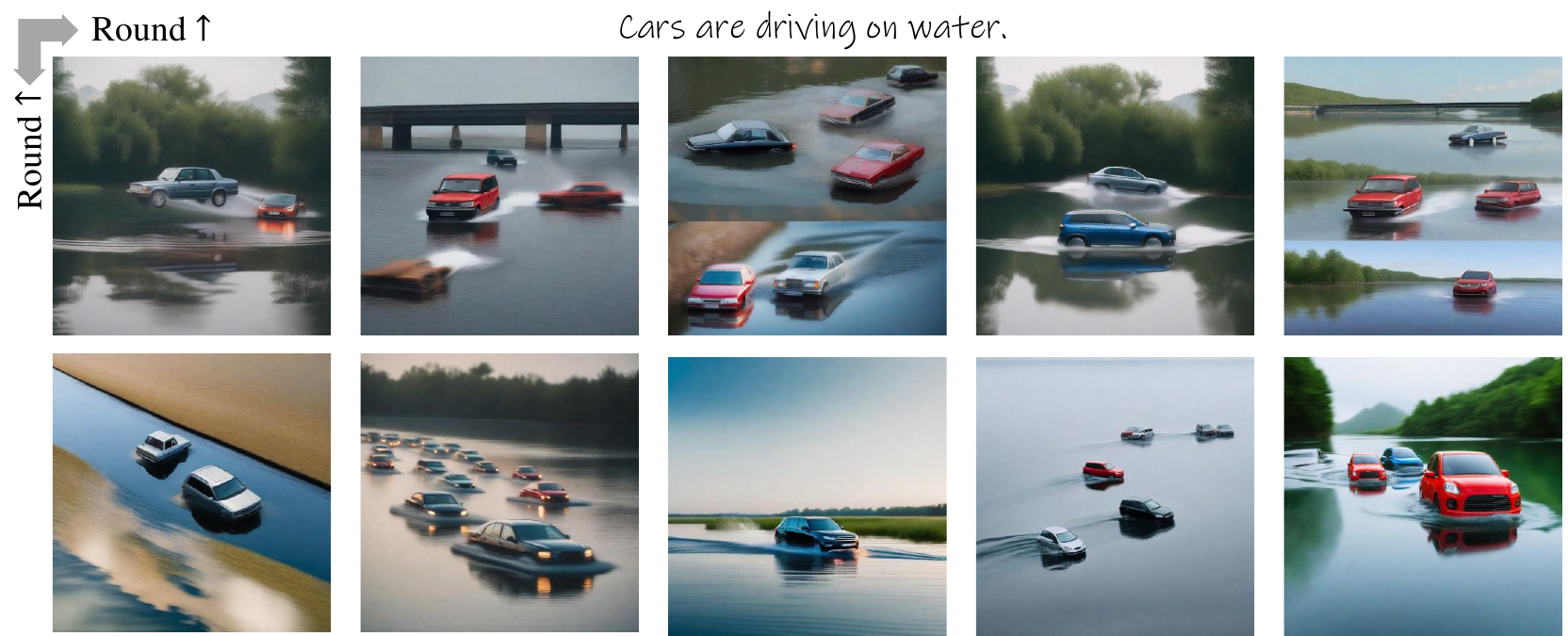}
    \caption{
     Reasoning generation process of prompt related to cars. Common sense dictates that cars drive on land, but the prompt asks for an image of a car on water. Initially, cars seems to fly out of the water, but in the final images, they drive across the center of the lake rather than float.
    } 
    \label{fig:reason_process_2}
\vspace{-4pt}
\end{figure}

Additional visualization of image reasoning process is shown in Fig.~\ref{fig:reason_process_2}. For results of the prompt in Fig.~\ref{fig:reason_process_2}, the general common sense is that cars drive on land, but the prompt requires generating an image of a car driving on water. In the initially generated images, the car appears to fly out of the water. In the final generated images, the car drives in the center of the lake rather than floats. 

\section{Prompt Details}\label{appen:prompt}

\subsection{Prompts Template "a(n) [animal] [activity]"}

\begin{figure}[h]
    \centering
    \includegraphics[width=1\textwidth]{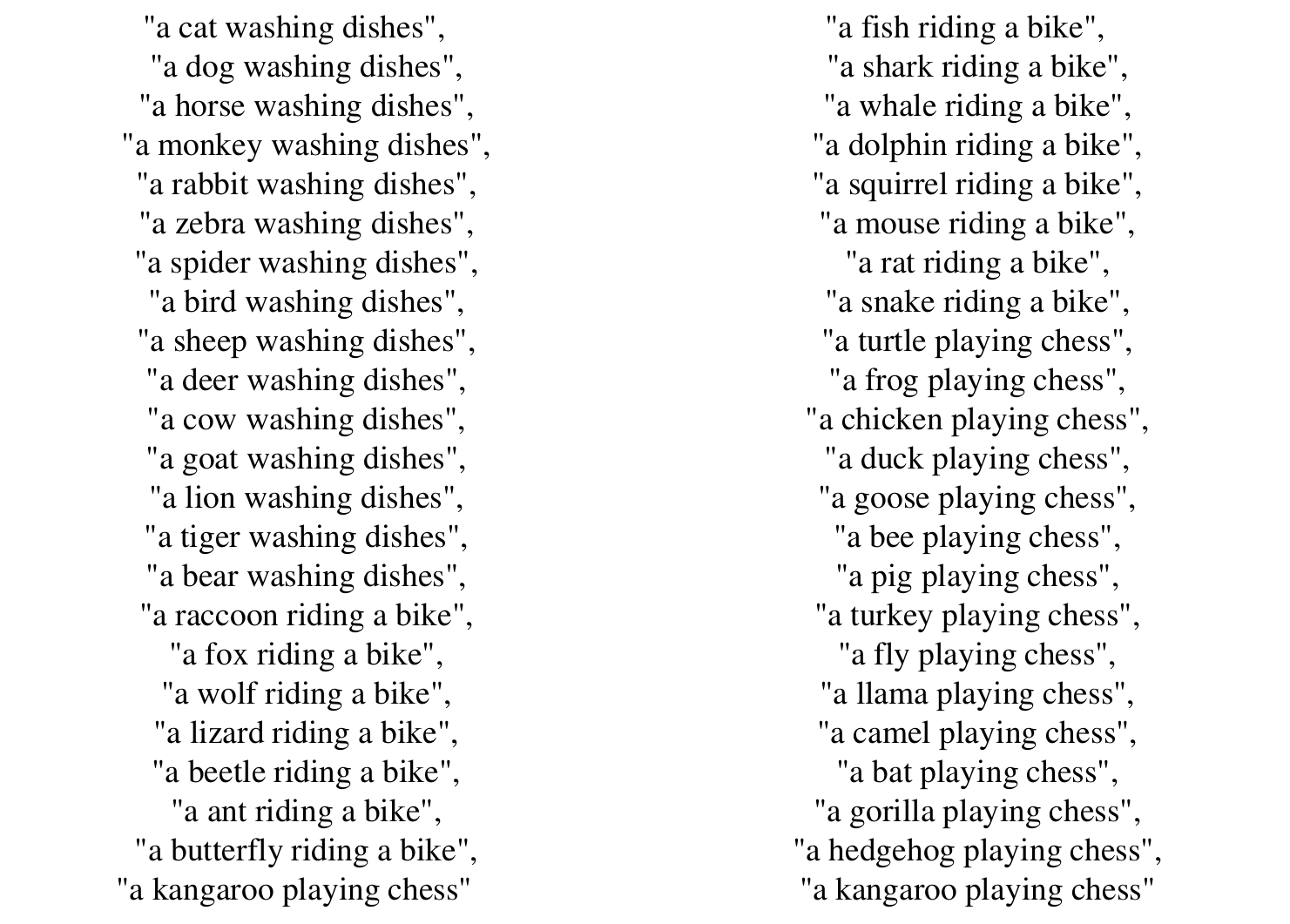}
    \caption{
    Prompts of the template "a(n) [animal] [activity]".
    } 
    \label{fig:appen_prompt_1}
\vspace{-4pt}
\end{figure}

Prompts of the template "a(n) [animal] [activity]" are shown in Fig.~\ref{fig:appen_prompt_1}, which are used to evaluate the text-image alignment of SRRL.

\subsection{Physical Phenomenon Related Prompts}

\begin{figure}[h]
    \centering
    \includegraphics[width=1\textwidth]{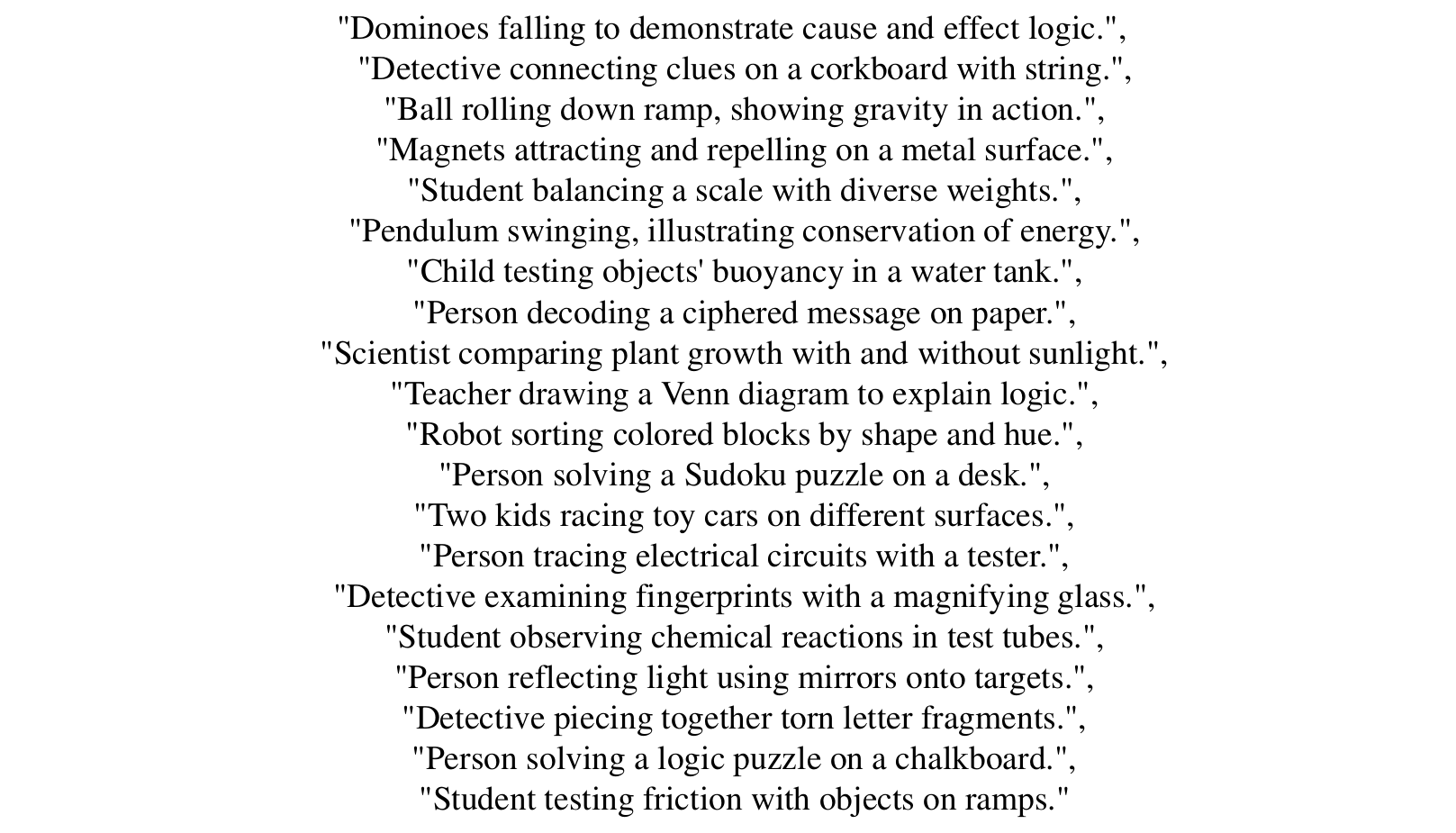}
    \caption{
    Physical phenomenon related prompts.
    } 
    \label{fig:appen_prompt_2}
\vspace{-4pt}
\end{figure}

Physical phenomenon related prompts are shown in Fig.~\ref{fig:appen_prompt_2}, which are used to evaluate the image reasoning ability of SRRL.

\subsection{Unconventional Physical Phenomena Prompts}

\begin{figure}[h]
    \centering
    \includegraphics[width=1\textwidth]{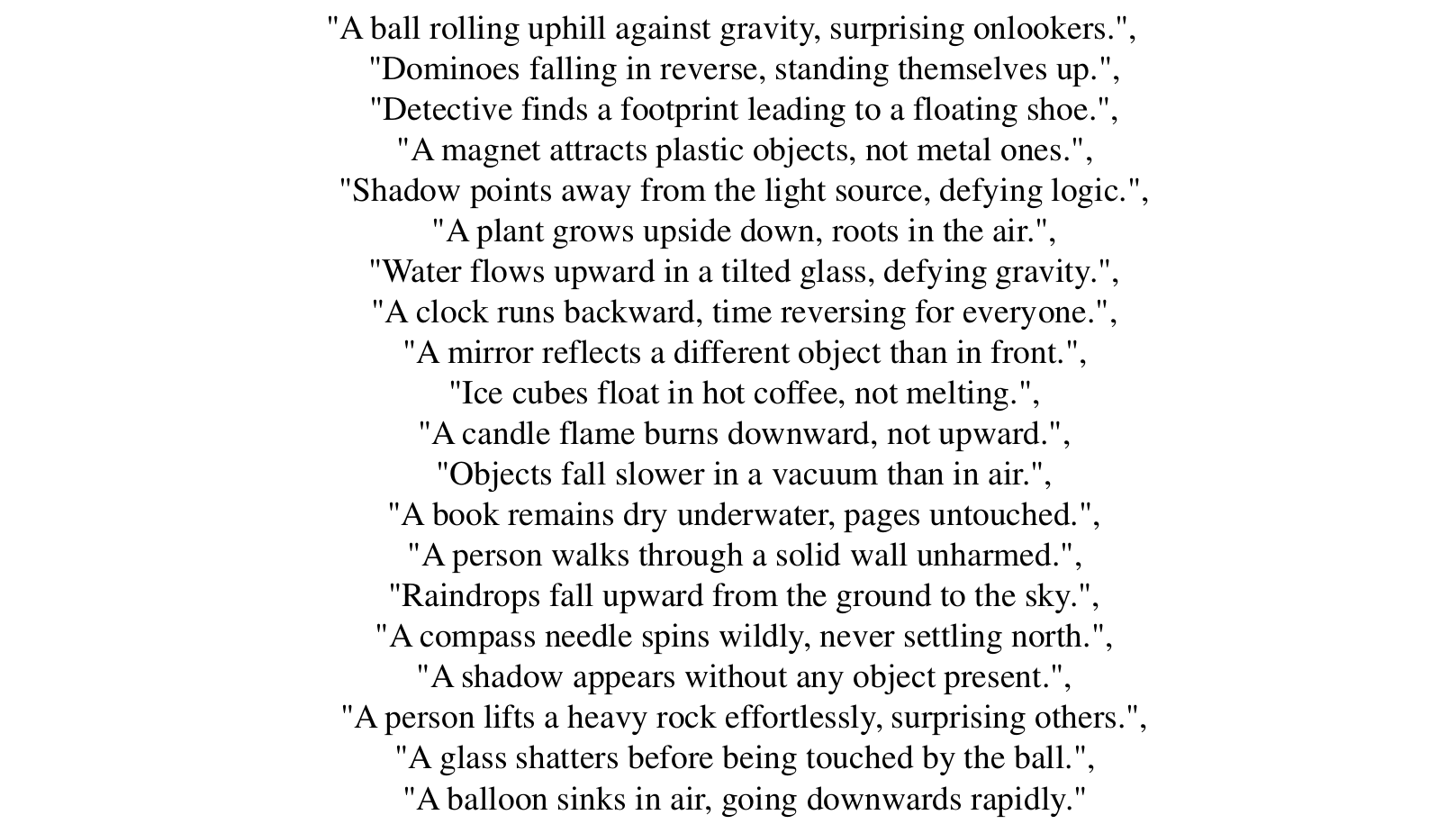}
    \caption{
    Unconventional physical phenomena prompts.
    } 
    \label{fig:appen_prompt_3}
\vspace{-4pt}
\end{figure}

Unconventional physical phenomena prompts are shown in Fig.~\ref{fig:appen_prompt_3}, which are used to evaluate the image reasoning ability of SRRL. 

Physical phenomenon related prompts and unconventional physical phenomena prompts are provided by GPT-4o. The prompts are: "Please help me think of some prompts generated from images that demonstrate logical reasoning, in English, and output in JSON format: [prompt1, prompt2,...]. Please provide me with 20 prompts" and "Please help me think of some prompts generated from images that demonstrate reasoning and unconventional phenomena. They should be in English and output in JSON format: [prompt1, prompt2,...]. Please provide me with 20 prompts.".

\section{Limitations}\label{appen:limitations}

We introduce three types of reward models, CLIP Score~\cite{hessel2021clipscore}, ImageReward~\cite{xu2023imagereward}, and VQAScore~\cite{lin2024evaluating} in the training process. However, these reward models are usually used to enhance text-image alignment or align with human feedback. Training a reward model of higher quality is of great significance for enhancing the reasoning ability of image generation models. We will reserve this for future work. Introducing better reward models can improve the accuracy of the reward function, leading to the broader application of reinforcement learning in image generation.

\section{Broader impacts}\label{appen:broader_impacts}

The advancement of image reasoning generation holds significant potential across various domains, including education, science, and creative industries. For education, image reasoning enables the creation of more sophisticated educational visuals, enhancing students' comprehension of scientific concepts. For science, image reasoning enables the creation of sophisticated research graphics, facilitating deeper comprehension of scientific progress. For creative industries, Image reasoning can generate intricate animated visuals, allowing the general public to experience the joy of animation.

\end{document}